\documentclass[10pt,twocolumn]{article}
\usepackage{arxiv}
\usepackage[T1]{fontenc}
\usepackage[utf8]{inputenc}
\usepackage{microtype}
\usepackage{amsmath,amssymb,amsfonts}
\usepackage{graphicx}
\usepackage{booktabs}
\usepackage{tabularx}
\usepackage{array}
\usepackage{multirow}
\usepackage{makecell}
\usepackage{float}
\usepackage[section]{placeins}
\usepackage{caption}
\usepackage{enumitem}
\usepackage{xcolor}
\usepackage[colorlinks=true,linkcolor=blue!55!black,citecolor=blue!55!black,urlcolor=blue!55!black]{hyperref}
\usepackage{url}

\hypersetup{pdftitle={SeLMRoute: Probabilistic Semantic Evidence for Large Language Model Routing},pdfauthor={Vasilis Perifanis},pdfkeywords={large language models, model routing, decision models, probabilistic representations, interpretable machine learning}}
\renewcommand{\headeright}{SeLMRoute preprint}
\renewcommand{\shorttitle}{SeLMRoute: Probabilistic Semantic Evidence for LLM Routing}
\title{SeLMRoute: Probabilistic Semantic Evidence for\\ Large Language Model Routing}

\author{
  \begin{tabular}[t]{c}
    \textbf{Vasilis Perifanis}\textsuperscript{1,2,3}\\
    \texttt{\small vasilis@indigma.eu}
  \end{tabular}
  \hspace{1.5em}
  \begin{tabular}[t]{c}
    \textbf{Nikolaos Pavlidis}\textsuperscript{1,2}\\
    \texttt{\small nikos@indigma.eu}
  \end{tabular}
  \hspace{1.5em}
  \begin{tabular}[t]{c}
    \textbf{Symeon Symeonidis}\textsuperscript{2}\\
    \texttt{\small ssymeoni@pme.duth.gr}
  \end{tabular}
  \\[1.5ex]
  {\footnotesize
  \textsuperscript{1}Indigma Innovations \quad
  \textsuperscript{2}Democritus University of Thrace \quad
  \textsuperscript{3}Athena Research Center}
}

\begin{document}

\twocolumn[{%
  \maketitle
}]

\begin{abstract}
Large language model (LLM) routing aims to select the most suitable model for each incoming query. Most existing routers learn this decision directly from query embeddings, model representations, preference data, or clusters of similar examples. Such approaches can be effective, yet the representation used for routing rarely states what a query actually requires. We introduce \textbf{SeLMRoute}, a routing framework that separates the extraction of candidate-independent semantic evidence from the learning of candidate performance and the application of deployment objectives. A decision model first evaluates a set of interpretable questions about the query, such as its reasoning requirements and use of external knowledge, with each judgment retained as a probability distribution. The resulting probabilistic semantic state is used by a lightweight supervised router to estimate candidate model performance. Routing objectives are applied after performance estimation, which allows the same semantic state to support performance-oriented and cost-aware decisions. On the LLMRouterBench (15 datasets, 20 candidate models, 11,481 queries), SeLMRoute achieves an average accuracy of $72.08\% \pm 0.45$, while grouped five-fold out-of-fold evaluation reaches $72.64\%$, compared with $69.23\%$ for the strongest fixed candidate. The representation achieves the highest mean performance among the evaluated semantic, dense, lexical, and domain-level representations. In a separate 13-model performance-cost setting, \textit{SeLMRoute} improves performance in all five grouped splits, with a mean PerfGain of $2.66\%$. Our code is available at \href{https://github.com/Indigma-Innovations/SeLMRoute}
{\textcolor{blue}{\texttt{https://github.com/Indigma-Innovations/SeLMRoute}}}.
\end{abstract}

\section{Introduction}\label{sec:intro}
Large language models (LLMs) exhibit substantial differences in their capabilities, computational requirements, and inference costs. A model that performs strongly on mathematical reasoning may be weaker on code, factual knowledge, instruction following, or affective tasks. Differences also remain among models of similar size because their data, training objectives, architectures, and specialization differ~\cite{li2026llmrouterbench}. A deployment system therefore faces a problem that is increasingly separate from language generation itself: \textit{Given a request and a set of available models, which model should receive the request?}

Model routing addresses this problem by learning a decision rule over a pool of candidate models. Earlier systems often focused on the trade-off between a strong expensive model and a weaker inexpensive model~\cite{chen2024frugalgpt,ding2024hybridllm,ong2025routellm}. Recent work considers larger model pools and richer representations. RouterDC learns query and model representations through contrastive objectives~\cite{chen2024routerdc}. EmbedLLM learns compact representations of model capability~\cite{zhuang2025embedllm}. GraphRouter represents queries, tasks, and models in a heterogeneous graph~\cite{feng2025graphrouter}. Model-SAT learns capability representations through aptitude-style instructions~\cite{zhang2025modelsat}. Avengers groups semantically similar requests before assigning them to models that performed well within each cluster~\cite{zhang2026avengers}. LLMRouterBench recently placed many of these approaches under a common evaluation framework and found that several leading routers obtain almost equivalent performance~\cite{li2026llmrouterbench}.

A common assumption connects many of these approaches. The router receives a representation of the query and learns how that representation relates to model performance. Dense embeddings are a natural choice because they provide a powerful summary of semantic similarity. A query about proving an inequality will usually be embedded near other mathematical questions. A programming request will usually be embedded near other programming requests. Similarity, however, is not the same as task requirement. Two queries may both concern Python while requiring very different capabilities. One may ask for the syntax of a dictionary comprehension. Another may require diagnosing a race condition across several interacting asynchronous functions under strict behavioral constraints. The distinction becomes clearer when the routing representation is made explicit. Consider the request: 
\begin{quote} 
\footnotesize 
``Implement a Python parser for the following configuration format. It must preserve comments, reject duplicate keys, report the exact line of a malformed entry, and remain compatible with the existing API.'' 
\end{quote}

A dense embedding represents the position of this request in the latent space and a classifier might label it as \emph{code}. However, a routing system would benefit from a richer description. The task requires code reasoning, high exactness, integration of several constraints, and some decomposition. Ambiguity is low and external factual knowledge is relatively unimportant. The requested behavior can be described through a small collection of semantic judgments whose meanings remain visible to a person inspecting the router. 

A second distinction concerns uncertainty. A judgment may not have one hard answer. A request can be partly mathematical and partly algorithmic, while decomposition may lie between ``moderate'' and ``high''. Hardening each judgment into its most likely value discards information before the routing problem has been learned. A probability distribution preserves the estimated semantic property and extractor's expressed uncertainty.

\textbf{SeLMRoute} is built around these two observations. The framework inserts an explicit \emph{probabilistic semantic state} between the raw query and the learned model router using a decision model that evaluates a fixed collection of questions about the request. Our implementation evaluates sixteen semantic dimensions covering reasoning type, knowledge requirements, task structure, ambiguity, exactness, and related properties. The raw probability mass returned by these judgments forms a 40-dimensional representation. A separate learner then estimates how well each candidate language model is expected to perform for that semantic state. 

The separation between semantic representation and performance learning reflects the different roles of the two components. The semantic extractor is independent of the set of candidate models and focuses on representing the characteristics of the incoming request. The performance learner then uses this representation to estimate how each candidate model is expected to behave for a given semantic state, without having to interpret the original text directly. These estimates are converted into a routing decision through a configurable objective. For instance, a performance-oriented deployment may select the model with the highest predicted quality, while a budget-sensitive deployment may jointly consider predicted quality and inference cost. Additional routing objectives can be introduced without modifying the semantic representation of the query.

Such a decomposition also creates an inspectable path from request to action. Suppose that a router selects a code-specialized model. A dense representation hardly explains the choice. \textit{SeLMRoute} can expose the evidence available to the routing learner, for example high probability mass on code reasoning, high exactness, high constraint density, and low dependence on current information. The learned mapping can still be nonlinear, and an explanation of the input evidence is not equivalent to a causal explanation of every tree decision. The representation nevertheless exposes substantially more structure than an anonymous embedding.

Recent benchmarks makes the representation question especially important. LLMRouterBench evaluates 10 routing approaches across 21 datasets and 33 models and reports a narrow band among several leading performance-oriented routers~\cite{li2026llmrouterbench}. On its reported performance-oriented evaluation, EmbedLLM reaches an AvgAcc of $71.24$, GraphRouter $70.29$, Model-SAT $71.88$, and Avengers $71.94$. LLMRouterBench argues that part of the gain may come from capturing coarse domain structure, since leading routers approach a Dataset Oracle that chooses one model per dataset. Such results raise a useful question: \textit{``Can a router describe a request through finer task requirements without returning to a large representation?''}.

Our experiments try to answer that question under duplicate-query-safe grouped evaluation. The probability-mass \textit{SeLMRoute} reaches $72.08\% \pm 0.45$ AvgAcc across five splits. The result is competitive with the strongest published LLMRouterBench routers, although published results use their own benchmark split protocol and are therefore treated as external comparisons. Within our controlled evaluation, probability mass performs above the full 88-feature semantic representation, hard semantic decisions, GTE-Qwen2 dense embeddings, coarse domain labels, and TF-IDF features. Grouped out-of-fold (OOF) evaluation reaches $72.64$, compared with $69.23$ for the best fixed candidate.

The architecture also survives changes in the semantic decision model. An open-weight implementation (Laya) reaches $70.53$ AvgAcc under the same grouped OOF protocol, which remains above the best fixed model. The JEV-based semantic state performs significantly better than the Laya-based state, indicating that semantic extractor quality matters even when the interface remains unchanged. A separate direct-routing experiment asks JEV to choose a model from anonymized candidate profiles. Its lower performance supports the architectural separation between semantic evidence extraction and downstream performance learning. Cost-aware routing presents a harder test. We evaluate 13 flagship models over 10 datasets and 12,446 benchmark instances using an inner validation set to select routing objectives before evaluation on the untouched test partition. \textit{SeLMRoute} produces positive PerfGain in all five grouped splits, with a mean improvement of $2.66\%$, but the strict protocol establishes no positive monetary savings. 

The main contributions of the paper are summarized as follows:
\begin{itemize} 
    \item We introduce an explicit probabilistic semantic state for LLM routing, where interpretable task requirements are estimated independently of the candidate model pool. 
    \item We separate semantic evidence extraction, model performance learning, and routing objectives, which permits the same query representation to support different model pools and deployment objectives. 
    \item We show that raw probability mass provides the strongest mean performance among the evaluated semantic representations and remains competitive with dense, lexical, and domain-level baselines. 
    \item We evaluate \textit{SeLMRoute} under grouped in-distribution splits, dataset-level and domain-level distribution shift, different routing learners, reduced semantic probe sets, an open-weight decision model, direct decision-model routing, and a separate performance-cost benchmark. 
\end{itemize}

The rest of this paper is organized as follows. Section~\ref{sec:related} reviews related work on LLM routing and query representations. Section~\ref{sec:framework} presents the SeLMRoute framework and discusses semantic evidence extraction, performance learning, and routing objectives. Section~\ref{sec:experiments} describes the experimental setup and evaluates routing performance, representation choices, generalization, cost-aware routing, and system overhead. Finally, Section~\ref{sec:conclusion} concludes the paper and outlines the directions for future work.

\section{Related Work}\label{sec:related}
\paragraph{Routing among language models.} 
External model routing differs from the routing used inside mixture-of-experts (MoE) architectures. Sparse MoE models learn gates that dispatch tokens or hidden states among internal expert networks~\cite{fedus2022switch}. LLM routing instead treats complete models as candidate systems and chooses which model should process an incoming request. Candidate models may have different architectures, training corpora, context limits, specializations, ownership, latency, and monetary cost. Such a setting permits routing across independently trained systems and does not require joint training of the candidate models. Early work emphasized inference cost. FrugalGPT studies adaptive model cascades and shows that requests can be escalated through models according to predicted utility~\cite{chen2024frugalgpt}. HybridLLM learns a router between a smaller and a larger model and allows the requested quality level to control the trade-off at inference time~\cite{ding2024hybridllm}. RouteLLM learns strong-versus-weak model routing from human preference data and studies transfer when the candidate pair changes~\cite{ong2025routellm}. The shared premise is that expensive capacity should be invoked when the request is expected to benefit from it. SeLMRoute retains a configurable objective layer, but its main contribution lies earlier in the pipeline, i.e., the query is first converted into a reusable semantic state before performance or cost enters the routing rule.

\paragraph{Query and model representations.} 
A second line of work focuses on the representation used to predict model suitability. RouterDC jointly learns query and model embeddings with sample-model and sample-sample contrastive losses~\cite{chen2024routerdc}. EmbedLLM learns compact model vectors intended to transfer across tasks such as routing and capability prediction~\cite{zhuang2025embedllm}. GraphRouter places query, task, and model nodes in a heterogeneous graph and learns model selection as an edge-prediction problem~\cite{feng2025graphrouter}. Model-SAT represents candidate capabilities through aptitude-style evaluations and trains a lightweight model to infer whether a candidate can handle an instruction~\cite{zhang2025modelsat}. ICL-Router develops in-context model representations that support the introduction of new candidates without retraining the complete router~\cite{wang2026iclrouter}. Each method addresses an important aspect of model selection, particularly the difficulty of representing heterogeneous candidate capabilities. Avengers takes a lightweight approach where queries are embedded and clustered, candidate models are scored within clusters, and new queries are assigned according to their nearest semantic cluster~\cite{zhang2026avengers}. Its strong benchmark performance is important because it shows that sophisticated routing architectures are not automatically superior to simple similarity structure. LLMRouterBench reaches a related conclusion under a unified evaluation, that several leading routers obtain close performance, while a substantial gap to the instance-level oracle remains~\cite{li2026llmrouterbench}. SeLMRoute focuses on a different representation question. The query vector is not intended to preserve general linguistic similarity. Each coordinate has a declared semantic interpretation before the performance learner is trained. A dimension can represent evidence about code reasoning, factual recall, exactness, ambiguity, constraint density, or another routing-relevant requirement. Probability distributions retain uncertainty within those judgments. Candidate model behavior is learned only after this semantic state has been constructed. The design therefore separates \emph{what the request appears to require} from \emph{which available model historically performs well under those requirements}.

Related interpretable routing work includes IRT-Router~\cite{song2025irt}, which uses item-response theory to model query attributes along with model abilities, and RADAR~\cite{fernandez2026radar}, which relates reasoning-task difficulty to model ability and reasoning budget. IRT-Router links its query and ability representations through an item-response formulation, while RADAR models task difficulty and reasoning-budget-dependent capability. SeLMRoute starts from sixteen human-readable typed questions and retains the decision model’s per-answer probability mass before learning candidate performance. Its forty features contain neither candidate identities nor a model-specific suitability score. A new model requires an empirical performance mapping, but does not require re-extracting the existing semantic vectors. 

\paragraph{Hard capability labels and probabilistic evidence.} 
Capability taxonomies provide a natural way to organize model behavior. Model-SAT explicitly constructs capability instructions~\cite{zhang2025modelsat}, while benchmark analyses frequently divide requests into domains such as mathematics, code, logic, knowledge, or affective reasoning~\cite{li2026llmrouterbench}. A domain label is useful but coarse. Fine-grained capability labels provide more structure, although a hard label still forces a semantic judgment into a single state. SeLMRoute instead retains atomic probability mass. For a score-type judgment, the representation contains the probability associated with the possible levels. For a categorical judgment, the representation retains probability assigned to the alternatives and a binary judgment similarly contributes its probability. The resulting representation resembles a small semantic belief state whose variables are defined by the system designer. Such a state permits uncertainty to influence the downstream learner without requiring the learner to reconstruct uncertainty from a dense text vector. Recent structured decision models provide a practical mechanism for producing such evidence. TypeSafe's JEV exposes typed primitives for binary, categorical, and ordinal decisions and returns probabilities together with the structured outputs~\cite{almeida2026jev,typesafe2026api}. SeLMRoute uses JEV as the primary semantic evidence extractor, but other open alternatives can be inserted. Note that a direct decision model can be asked ``which candidate should answer this query?'' In SeLMRoute, however, the decision model is asked a set of reusable questions about the query, while observed candidate outcomes train the final mapping from semantic state to model utility. Our direct-routing ablation evaluates the difference experimentally.

\paragraph{Benchmarking model routers.} 
Comparisons across routing papers have historically been difficult since studies use different candidate pools, datasets, costs, metrics, and model outputs. LLMRouterBench provides a common evaluation framework with more than 400,000 collected model responses across 21 datasets and 33 models, along with performance-oriented and performance-cost settings and adapters for representative routers~\cite{li2026llmrouterbench}. Its performance-oriented pool contains 20 lightweight models, while its performance-cost pool contains 13 flagship models. The benchmark also defines AvgAcc, Gain@R, Gain@B, and Gap@O, which permit routers to be compared against random selection, a best fixed model, and an instance-level oracle. Our evaluation follows the LLMRouterBench model pools and metrics while adding safeguards required by a semantic router. Identical normalized router inputs are kept within the same learned-router partition, which prevents repeated queries from appearing on both sides of an in-distribution split. Statistical comparisons group duplicate semantic inputs during resampling. Cost-aware operating points are selected exclusively on an inner validation partition before final test evaluation. Published LLMRouterBench systems are reported as external reference points, while claims of paired improvement are restricted to systems evaluated under our shared grouped protocol. The resulting evaluation asks a broader question than whether one router obtains the highest benchmark score. The central question is whether model routing benefits from an intermediate representation whose features remain understandable, whose uncertainty is preserved, and whose meaning does not depend on the identities of the models being routed.

\section{The SeLMRoute Framework}
\label{sec:framework}
\begin{figure*}[ht] \centering \includegraphics[width=0.8\textwidth]{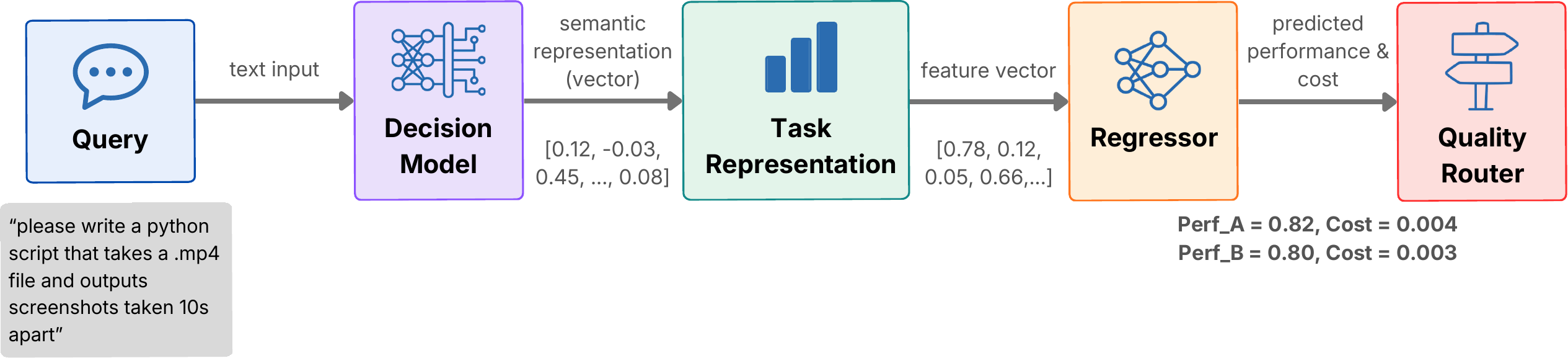} 
    \caption{Overview of SeLMRoute. A query is first transformed into an interpretable probabilistic semantic state by a typed decision model. A learned performance model maps the semantic state to predicted candidate model performance. The routing objective then selects a model according to user-defined preferences such as predicted quality or a quality-cost trade-off.} 
    \label{fig:SeLMRoute_architecture} 
\end{figure*}

SeLMRoute separates three questions that are often represented into a single routing model: what a request requires, how candidate models are expected to perform on such a request, and which trade-off the deployment wishes to optimize. Figure~\ref{fig:SeLMRoute_architecture} summarizes the resulting pipeline. An input query is first converted into a probabilistic semantic state through a set of typed human-readable judgments. A learned performance model maps that state to an expected score for every candidate model. A routing policy then converts the predicted scores, and optionally predicted costs or other deployment constraints, into a final model choice.

The separation is useful because the three components solve different problems. Semantic evidence describes the request and has meaning independently of the candidate pool. Performance learning captures empirical differences among the available models, while the routing objective expresses a deployment preference. A change in price, quality preference, or candidate availability therefore need not redefine the semantic evidence. Adding a previously unevaluated model does require candidate-performance evidence or model-specific calibration.
\subsection{Problem Formulation}
\label{sec:problem_formulation}

Let $x \in \mathcal{X}$ denote an input query and let $\mathcal{M} = \{m_1,m_2,\ldots,m_K\}$ be a pool of $K$ candidate language models. For each query-model pair $(x,m)$, let $y_m(x) \in [0,1]$ denote the observed task score of model $m$ on query $x$. The score can represent binary correctness, pass rate, success rate, or another benchmark-specific measure that has been normalized to the same direction, where larger values are preferred. A cost-aware setting additionally associates the pair with a non-negative inference cost $c_m(x) \geq 0$.

A router is a policy
\begin{equation}
    r : \mathcal{X} \rightarrow \mathcal{M}
\end{equation}
that selects one candidate for each query. A purely performance-oriented router seeks a model with high expected $y_m(x)$. A deployment with economic or operational constraints may instead optimize a utility function that combines expected performance with cost, latency, availability, or another measurable property.

Historical routing data can be represented as $\mathcal{D} = \left\{\left(x_i,\mathbf{y}_i,\mathbf{c}_i\right)\right\}_{i=1}^{N}$, where $\mathbf{y}_i = \left[y_{i1},\ldots,y_{iK}\right]$ contains the observed candidate scores for query $x_i$, and $\mathbf{c}_i$ contains the corresponding costs when cost information is available. Some candidate-query outcomes may be missing. We represent their availability with $a_{im}\in\{0,1\}$.

A supervised router can learn the model decision directly from a textual representation of $x$. SeLMRoute instead factorizes the routing function as

\begin{equation}
    x
    \xrightarrow{\;S\;}
    \mathbf{s}(x)
    \xrightarrow{\;F_{\theta}\;}
    \widehat{\mathbf{y}}(x)
    \xrightarrow{\;R_{\omega}\;}
    r(x),
    \label{eq:SeLMRoute_factorization}
\end{equation}
where $S$ extracts semantic evidence, $F_{\theta}$ predicts candidate performance, and $R_{\omega}$ applies the user-selected routing objective. The parameter $\omega$ describes deployment preferences and is kept separate from the parameters $\theta$ learned from historical candidate performance.

Equation~\ref{eq:SeLMRoute_factorization} is the central design choice of SeLMRoute. The intermediate state $\mathbf{s}(x)$ is defined before candidate performance is considered, hence, its dimensions describe the input query.

\subsection{Architecture and Semantic Evidence Extraction}
\label{sec:semantic_extraction}

The semantic extractor evaluates a collection of typed questions $\mathcal{Q} = \{q_1,q_2,\ldots,q_J\}$. Each question $q_j$ has a declared answer space $\mathcal{A}_j$. A decision model $D$ receives the query and returns a probability distribution:

\begin{equation}
    \mathbf{p}_j(x)
    =
    \left[
        p_j(a_1 \mid x),
        \ldots,
        p_j(a_{|\mathcal{A}_j|} \mid x)
    \right]
    \in
    \Delta^{|\mathcal{A}_j|-1},
    \label{eq:probe_distribution}
\end{equation}
where $\Delta$ denotes the probability simplex.

Our implementation uses JEV as the primary decision model. JEV exposes typed Noul, Choice, and Score decisions with probabilistic outputs~\cite{almeida2026jev,typesafe2026api}. Nevertheless, SeLMRoute does not depend on JEV-specific model internals. Any backend that can answer the same typed questions and return compatible probabilities can replace it. Section~\ref{sec:experiments} evaluates this property with the open-weight Laya model~\cite{nandhakishor2026laya}.

The deployed semantic schema contains sixteen fine-grained probes. Eight are binary Noul questions and eight are four-level Score questions. A seventeenth Choice question estimates a coarse task family for diagnostics. The task-family output is excluded from the main SeLMRoute representation. This way, the learned router receives the fine-grained semantic evidence and not an explicit mathematics, code, logic, or knowledge label. Table~\ref{tab:semantic_probes} lists the sixteen routing probes.

\begin{table*}[t]
\centering
\small
\caption{Fine-grained semantic probes used by SeLMRoute. Noul probes return a probability for a binary judgment. Score probes return probability mass over four ordered levels.}
\label{tab:semantic_probes}
\begin{tabular}{p{0.22\textwidth} p{0.10\textwidth} p{0.58\textwidth}}
\toprule
\textbf{Probe} & \textbf{Type} & \textbf{Semantic question} \\
\midrule
\texttt{math\_reasoning}
& Noul
& Whether correctness requires mathematical calculation, symbolic manipulation, or quantitative reasoning. \\

\texttt{code\_reasoning}
& Noul
& Whether correctness requires writing, modifying, debugging, or reasoning about executable code. \\

\texttt{formal\_logic}
& Noul
& Whether the task materially requires formal, combinatorial, rule-based, or constraint-satisfaction reasoning. \\

\texttt{factual\_recall}
& Noul
& Whether the request can be solved primarily through factual knowledge with little derivation. \\

\texttt{social\_affective}
& Noul
& Whether correctness depends on emotion, intention, interpersonal meaning, or social context. \\

\texttt{tool\_interaction}
& Noul
& Whether completing the task requires interaction with an external tool, API, software environment, or simulator. \\

\texttt{external\_knowledge}
& Noul
& Whether required factual information is absent from the supplied query and cannot be derived from it. \\

\texttt{current\_information}
& Noul
& Whether correctness depends materially on recent or changing information. \\

\texttt{domain\_specialization}
& Score
& The degree of specialized disciplinary knowledge required. \\

\texttt{reasoning\_depth}
& Score
& The number and dependence of reasoning steps needed for a correct answer. \\

\texttt{constraint\_density}
& Score
& The number of independent requirements that must be satisfied simultaneously. \\

\texttt{context\_integration}
& Score
& The amount of information from separate parts of the query that must be combined. \\

\texttt{decomposition\_need}
& Score
& The extent to which the task should be separated into intermediate subproblems. \\

\texttt{ambiguity}
& Score
& The degree to which the intended task or its success condition is underspecified. \\

\texttt{exactness}
& Score
& The sensitivity of correctness to precise values, details, constraints, or output behavior. \\

\texttt{answer\_openness}
& Score
& The breadth of substantively different answers that could reasonably be correct. \\
\bottomrule
\end{tabular}
\end{table*}

The questions are phrased as task properties, with no probe asking whether a particular candidate is suitable and no candidate name appears in the semantic schema. The decision model therefore has no direct mechanism for learning that a particular semantic pattern should be mapped to a particular candidate.

A single query can activate several semantic dimensions at once. Consider the running example
\begin{quote}
    \footnotesize
    ``Please write a Python script that takes a \texttt{.mp4} file and outputs screenshots taken 10 seconds apart.''
\end{quote}
The query has strong evidence for code reasoning and requires some external programming knowledge. Exactness is crucial because the temporal interval and file behavior must be implemented correctly. Formal logic and social-affective interpretation, on the other hand, have little relevance. Tool interaction has a different meaning from code reasoning in our schema. Specifically, the requested answer is code, while the language model is not itself required to execute a video-processing tool. Such distinctions are difficult to express through a single domain label such as \emph{code}.

All fine-grained questions are evaluated against the same query and their outputs are then converted into a fixed-length semantic state. The decision model performs semantic analysis once, after which different candidate pools and routing objectives can reuse the result.

\subsection{Interpretable and Hard Semantic Judgments}
\label{sec:semantic_representation}

SeLMRoute retains the probability mass produced by the semantic extractor instead of reducing every judgment to its most likely answer. 

Let $\mathcal{N}$ denote the eight binary Noul probes and $\mathcal{S}$ denote the eight four-level Score probes. A Noul probe is represented by its probability of a positive judgment, $p_j(x) = P(q_j=\mathrm{yes}\mid x), j \in \mathcal{N}$. The complementary probability is determined by $1-p_j(x)$ and therefore does not require a second feature. A Score probe returns probability mass over four ordered levels, $\mathbf{p}_j(x) = \left[p_{j1}(x), p_{j2}(x), p_{j3}(x), p_{j4}(x) \right], j \in \mathcal{S}$.

The probability-mass representation used by the main SeLMRoute router is:

\begin{equation}
\begin{split}
    \mathbf{s}_{\mathrm{PM}}(x)
    =
    \Big[
        &\{p_j(x)\}_{j\in\mathcal{N}},\\
        &\{\mathbf{p}_j(x)\}_{j\in\mathcal{S}}
    \Big].
\end{split}
\end{equation}

Its dimensionality follows directly from the schema:
\begin{equation}
    d_{\mathrm{PM}}
    =
    8 + 8 \times 4
    =
    40.
    \label{eq:pm_dim}
\end{equation}

The representation preserves a distinction that a hard label removes. Suppose a Score probe has probability mass $[0.05,\;0.20,\;0.55,\;0.20]$. A hard decision records level 3. The probability-mass state additionally records that level 2 and level 4 remain plausible. A second query with distribution $[0.00,\;0.01,\;0.98,\;0.01]$ receives the same hard label despite presenting much stronger semantic evidence for that level. SeLMRoute allows the downstream performance learner to distinguish the two cases.

The corresponding hard representation contains one scalar per probe. For Noul decisions, $h_j(x) = \mathbb{I} \left[p_j(x) \geq 0.5 \right], j \in \mathcal{N}$, while a Score decision becomes $h_j(x) = \arg\max_{k\in\{1,2,3,4\}} p_{jk}(x), j \in \mathcal{S}$. The resulting hard state has $d_{\mathrm{hard}} = 16$ features. Section~\ref{sec:experiments} compares this state directly with the probability-mass representation under identical routing models and data splits.

The decision outputs can also support derived statistics. For a probe with $L_j$ possible outcomes, the normalized semantic entropy is:
\begin{equation}
    H_j(x)
    =
    -
    \frac{
        \sum_{k=1}^{L_j}
        p_{jk}(x)\log p_{jk}(x)
    }{
        \log L_j
    }.
    \label{eq:semantic_entropy}
\end{equation}

A value near zero indicates a concentrated semantic judgment, while a value near one indicates a diffuse distribution. We use these entropy values in later uncertainty experiments.

Our full semantic representation contains the probability mass together with the decision model's derived quantities such as entropy, confidence, expected score, and distributional spread. It contains 88 fine-grained semantic features in the current schema. The larger representation can be used for the learning, but it is not the default SeLMRoute state. The main system uses the 40 raw probability-mass features, as defined in Equation~\ref{eq:pm_dim}.

Interpretability in SeLMRoute refers to the semantic evidence supplied to the performance learner. Each input dimension has a declared meaning and can be inspected independently. Note that the representation does not provide a complete explanation of the downstream regressor. A nonlinear performance model can combine several semantic signals in ways that are more complex than a rule list. The important property is that the information available to that model remains explicit.

\subsection{Learning Model Performance}
\label{sec:performance_learning}

The semantic state does not select a candidate directly. A supervised performance model learns the relationship between semantic requirements and observed candidate behavior. Let $F_{\theta} : \mathbb{R}^{d} \rightarrow \mathbb{R}^{K}$ denote the performance predictor. For a semantic state $\mathbf{s}(x)$,

\begin{equation}
    F_{\theta}(\mathbf{s}(x))
    =
    \widehat{\mathbf{y}}(x)
    =
    \left[
        \hat{y}_1(x),
        \ldots,
        \hat{y}_K(x)
    \right],
    \label{eq:performance_vector}
\end{equation}
where $\hat{y}_m(x)$ estimates the expected benchmark score of candidate $m$ on the query.

The training objective can be written as:

\begin{equation}
    \mathcal{L}_{\mathrm{perf}}(\theta)
    =
    \frac{
        \sum_{i=1}^{N}
        \sum_{m=1}^{K}
        a_{im}
        \ell
        \left(
            y_{im},
            \hat{y}_{im}
        \right)
    }{
        \sum_{i=1}^{N}
        \sum_{m=1}^{K}
        a_{im}
    },
    \label{eq:performance_loss}
\end{equation}
where $a_{im}$ masks unavailable outcomes. Squared error is used in our implementation.

The primary SeLMRoute implementation uses a multi-output CatBoost regressor~\cite{prokhorenkova2018catboost}. One model jointly predicts the score vector in Equation~\ref{eq:performance_vector} for the rectangular performance-oriented candidate pool. The output space corresponds to candidate models, while the input contains only semantic evidence about the query.

The larger performance-cost benchmark contains missing candidate-query cells. Candidate-specific regressors are used there so that each model can be trained from its observed outcomes without score imputation. The distinction affects model fitting, but it does not alter the semantic representation or the routing objective.

A useful consequence follows from the factorization. Model identities are absent from $\mathbf{s}(x)$. A change in the candidate pool therefore does not require semantic evidence to be regenerated for queries that have already been encoded. The performance mapping must still be updated when a new model is introduced because the system needs empirical evidence about that model. SeLMRoute separates that empirical update from query understanding.

Consider again the video-processing request. Suppose two available candidates receive predicted scores $\hat{y}_A(x)=0.82$ and $\hat{y}_B(x)=0.80$. A performance-only router selects $m_A$ with no semantic probe containing the rule ``choose model A.'' The preference arises from historical evidence that model $A$ performs slightly better on requests with similar semantic states.

\subsection{User-Configurable Routing Objectives}
\label{sec:routing_objectives}

Predicted performance and routing preference are kept separate. A performance-oriented deployment uses
\begin{equation}
    r_{\mathrm{quality}}(x)
    =
    \arg\max_{m\in\mathcal{M}}
    \hat{y}_m(x).
    \label{eq:quality_router}
\end{equation}

Equation~\ref{eq:quality_router} is the default policy in the main performance experiments. A cost-sensitive deployment additionally learns or supplies an expected candidate cost $\hat{c}_m(x)$.

Quality and cost occupy different numerical scales, so our implementation normalizes both quantities using ranges estimated from the fitting partition. Let

\begin{equation}
    \widetilde{y}_m(x)
    =
    \frac{
        \hat{y}_m(x)-y_{\min}
    }{
        y_{\max}-y_{\min}+\epsilon
    }
\end{equation}

and

\begin{equation}
    \widetilde{c}_m(x)
    =
    \frac{
        \hat{c}_m(x)-c_{\min}
    }{
        c_{\max}-c_{\min}+\epsilon
    }.
\end{equation}

The cost-aware utility is then

\begin{equation}
    U_m(x;\lambda)
    =
    (1-\lambda)
    \widetilde{y}_m(x)
    -
    \lambda
    \widetilde{c}_m(x),
    \lambda\in[0,1],
    \label{eq:cost_utility}
\end{equation}

and the corresponding route is

\begin{equation}
    r_{\lambda}(x)
    =
    \arg\max_{m\in\mathcal{M}}
    U_m(x;\lambda).
    \label{eq:cost_route}
\end{equation}

The case $\lambda=0$ reduces to predicted-quality maximization after normalization, where larger values place greater weight on cost. In our performance-cost experiments, $\lambda$ is selected on an inner validation partition and is fixed before the final test evaluation. No test score is used to choose the reported operating point. The resulting evaluation follows the performance-cost setting of LLMRouterBench~\cite{li2026llmrouterbench}.

A deployment can also restrict the candidate set before utility maximization. Let $\mathcal{M}_{\pi}(x) \subseteq \mathcal{M}$ denote candidates permitted by a user-defined policy $\pi$. Such a policy can remove unavailable models or enforce deployment rules before the learned utility is considered. The generic decision becomes

\begin{equation}
    r(x)
    =
    \arg\max_{m\in\mathcal{M}_{\pi}(x)}
    U_m(x;\omega).
    \label{eq:general_route}
\end{equation}
Our experiments instantiate quality-only routing and the quality-cost objective in Equation~\ref{eq:cost_utility}. 

The separation between semantic evidence and routing objectives is also what distinguishes SeLMRoute from direct decision-model routing. A direct approach can present candidate profiles to a decision backend and ask it to choose a model in one step. SeLMRoute instead asks the backend to describe the request, then learns the relationship between that description and measured candidate outcomes. The direct-routing baseline in Section~\ref{sec:experiments} evaluates whether the intermediate semantic state contributes beyond the decision backend itself.

\section{Experiments}
\label{sec:experiments}
Our experiments try to answer the following research questions:
\begin{enumerate}[label=\textbf{RQ\arabic*:}, leftmargin=*]
    \item Can an explicit semantic state route models competitively compared with other routing approaches?
    \item Does the representation of semantic evidence affect routing performance?
    \item Does the separation between semantic analysis and model performance learning contribute beyond direct decision-model routing?
    \item How well does the semantic state transfer across datasets, domains, decision models, and reduced probe sets?
    \item Can the same semantic representation support performance-cost routing without changing the semantic extractor?
\end{enumerate}

\subsection{Experimental Setup}
\label{sec:experimental_setup}

\paragraph{Benchmark.}
We use LLMRouterBench~\cite{li2026llmrouterbench}, which provides frozen model responses and per-instance scores for unified routing evaluation. The main performance-oriented setting contains 20 lightweight candidate models evaluated across 15 datasets. Our frozen benchmark bundle contains 11,481 query instances and 229,620 candidate outcomes. The LLMRouterBench paper reports 11,480 prompts in this setting. The one-instance discrepancy comes from the frozen MMLU-Pro result bundle, which contains 1,001 records while the published benchmark count is 1,000. All internal SeLMRoute comparisons use exactly the same 11,481-instance support.

\begin{table}[t]
    \centering
    \small
    \caption{Performance-oriented benchmark used for the main SeLMRoute evaluation.}
    \label{tab:benchmark_datasets}
    \begin{tabular}{lr}
    \toprule
    \textbf{Dataset} & \textbf{Queries} \\
    \midrule
    AIME & 60 \\
    BBH & 1,080 \\
    EmoryNLP & 697 \\
    FinQA & 1,147 \\
    GPQA & 198 \\
    HumanEval & 164 \\
    K\&K & 700 \\
    KorBench & 1,250 \\
    LiveCodeBench & 1,055 \\
    MATH-500 & 500 \\
    MathBench & 150 \\
    MBPP & 974 \\
    MedQA & 1,273 \\
    MELD & 1,232 \\
    MMLU-Pro & 1,001 \\
    \midrule
    \textbf{Total} & \textbf{11,481} \\
    \bottomrule
\end{tabular}
\end{table}

The datasets cover mathematics, code, formal and commonsense reasoning, factual knowledge, medicine, finance, and affective understanding. Candidate models follow the performance-oriented LLMRouterBench pool. The strongest fixed candidate across the grouped OOF evaluation is Qwen3-8B. Table~\ref{tab:benchmark_datasets} summarizes the considered datasets.

\paragraph{Semantic extraction.}
JEV serves as the primary decision model, with each query evaluated with the sixteen probes defined in Table~\ref{tab:semantic_probes}. The default \textit{ProbabilityMass representation} contains 40 features, \textit{Full representation} contains 88 semantic features, including probability mass and derived statistics, and \textit{Hard representation} contains 16 discrete semantic judgments. Semantic extraction is performed once per query and reused across all downstream router experiments.

A separate experiment replaces JEV with the open-weight Laya decision model, while the probe schema and downstream routing procedure remain unchanged. The experiment tests whether the SeLMRoute architecture depends on a particular semantic extractor.

\paragraph{Dense and lexical baselines.}
We compare SeLMRoute with three alternative query representations. \textbf{GTE-Qwen2}\footnote{\url{https://huggingface.co/Alibaba-NLP/gte-Qwen2-7B-instruct}}  is a large instruction-tuned text embedding model~\cite{li2023towards}  that captures general semantic information.  \textbf{TF-IDF} represents queries using lexical term weights, while  \textbf{DomainOnly} uses only the broad task domain assigned to each  benchmark dataset, namely mathematics, code, logic, knowledge, or  affective tasks. Unlike the other representations, DomainOnly does  not use the query text or distinguish between individual queries within the same domain and provides a baseline for assessing how much routing performance can be explained by general  task categories alone. All three representations are evaluated using the same downstream regression procedure,  allowing us to examine how the information provided to the router affects its ability to predict candidate model performance.

\paragraph{Grouped train-test protocol.}
To avoid data leakage, we use a grouped evaluation protocol that keeps repeated queries within the same data partition. LLMRouterBench contains instances with identical query text but different record identifiers. If these instances are randomly assigned to the train and test sets, the performance learner may encounter a query during training that also appears in the test set. Such overlap could lead to overly optimistic performance estimates by allowing the router to learn from repeated inputs.
\begin{table*}[t]
    \centering
    \small
    \caption{Grouped performance-oriented evaluation. Results are mean $\pm$ standard deviation over five 70/30 grouped splits.}
    \label{tab:main_results}
    \begin{tabular}{lcccc}
    \toprule
    \textbf{Method} &
    \textbf{AvgAcc} &
    \textbf{Gain@R (\%)} &
    \textbf{Gain@B (\%)} &
    \textbf{Gap@O (\%)} \\
    \midrule
    \textbf{SeLMRoute ProbabilityMass}
    & $\mathbf{72.08\pm0.45}$
    & $\mathbf{56.08\pm1.62}$
    & $\mathbf{5.72\pm0.67}$
    & $\mathbf{22.00\pm0.67}$ \\
    
    SeLMRoute Full
    & $71.30\pm0.18$
    & $54.11\pm1.40$
    & $4.58\pm0.52$
    & $22.84\pm0.44$ \\
    
    SeLMRoute Hard
    & $71.20\pm0.59$
    & $54.61\pm2.28$
    & $4.41\pm1.11$
    & $22.94\pm0.87$ \\
    
    GTE-Qwen2 + CatBoost
    & $71.16\pm0.74$
    & $53.32\pm2.71$
    & $4.34\pm0.93$
    & $23.01\pm0.90$ \\
    
    DomainOnly
    & $71.08\pm1.03$
    & $53.45\pm2.19$
    & $4.45\pm0.82$
    & $23.16\pm1.13$ \\
    
    TF-IDF
    & $70.87\pm0.78$
    & $53.25\pm1.18$
    & $3.82\pm1.05$
    & $23.29\pm1.25$ \\
    \bottomrule
\end{tabular}
\end{table*}

We address this by grouping instances according to their dataset and normalized query text, i.e., $g(x)=\left(d(x),\operatorname{norm}(x)\right),$ where $d(x)$ is the source dataset and $\operatorname{norm}(x)$ denotes the normalized query supplied to the router. Instances that belong to the same dataset and share the same normalized query are assigned to a single group, regardless of their original record identifiers. The benchmark contains 11,481 instances, corresponding to 11,423 unique groups. Among these, 47 groups contain repeated queries, accounting for 58 additional instances. All instances within a group are assigned to the same partition, ensuring that the same normalized query from a given dataset cannot appear in both training and testing.

For the main evaluation, we independently partition each dataset into 70\% training and 30\% test instances, while preserving the groups defined above. We repeat this procedure using five random seeds, $\{42,999,2024,2025,3407\}$, producing five distinct train-test splits. Each representation is evaluated on the same splits, allowing direct comparisons under identical training and testing conditions. We report the mean and standard deviation across these five runs in the main performance table.

We additionally perform a five-fold grouped OOF evaluation. The data are divided into five folds, with all instances from the same group assigned to the same fold. In each iteration, the learner is trained on four folds and evaluated on the remaining fold. After five iterations, every benchmark instance has received a prediction from a model that was not trained on its group. Unlike the repeated 70/30 evaluation, this process provides held-out predictions for the entire benchmark. We use these predictions for paired representation comparisons, decision model comparisons, and the nested probe-reduction experiment. The grouped OOF predictions also form the basis of our paired statistical tests.

\paragraph{Metrics.}
We follow the performance metrics defined in LLMRouterBench~\cite{li2026llmrouterbench}. Let $\operatorname{Acc}(a,d)$ denote the accuracy of router $a$ on dataset $d$. The average accuracy is:
\begin{equation}
    \operatorname{AvgAcc}(a)
    =
    \frac{1}{|\mathcal{D}|}
    \sum_{d\in\mathcal{D}}
    \operatorname{Acc}(a,d).
\end{equation}

We compare SeLMRoute against three reference baselines defined by LLMRouterBench~\cite{li2026llmrouterbench}. \textbf{Random Router} ($R$) selects a candidate uniformly at random for each query. \textbf{Best Single} ($B$) identifies the individual model with the highest average accuracy across all evaluation datasets and uses that same model for every query.  \textbf{Oracle} ($O$) uses the observed candidate outcomes to select a correct model for each query, whenever one exists. Unlike the learned routers, the Oracle has access to the actual outcomes and represents a theoretical upper bound on routing performance.

The relative gains are:

\begin{equation}
    \operatorname{Gain@}b
    =
    \frac{1}{|\mathcal{D}|}
    \sum_{d\in\mathcal{D}}
    \left(
        \frac{\operatorname{Acc}(a,d)}
             {\operatorname{Acc}(b,d)}
        -1
    \right),
    b\in\{R,B\},
\end{equation}

and the oracle gap is:

\begin{equation}
    \operatorname{Gap@O}
    =
    \frac{1}{|\mathcal{D}|}
    \sum_{d\in\mathcal{D}}
    \left(
        1-
        \frac{\operatorname{Acc}(a,d)}
             {\operatorname{Acc}(O,d)}
    \right).
\end{equation}

Gain and gap values are reported as percentages.

\paragraph{Statistical tests.}
We use the grouped OOF predictions for statistical comparisons, ensuring that competing methods are evaluated on the same held-out queries. To account for uncertainty, we compute 95\% bootstrap confidence intervals by resampling query groups within each dataset and recalculating the difference in average accuracy between methods. Statistical significance is assessed using two-sided paired permutation tests, which randomly reverse the paired performance differences at the group level to test whether the observed difference could arise by chance.

Both procedures preserve the grouping of repeated queries, preventing identical inputs from being treated as independent observations. Dataset-aware resampling also preserves the contribution of each dataset to the macro-averaged accuracy, using 10,000 bootstrap replications and 20,000 permutations.

\subsection{Main Performance-Oriented Routing}
\label{sec:main_performance}

Table~\ref{tab:main_results} presents the results across the five grouped train-test splits. ProbabilityMass achieves the highest mean AvgAcc of $72.08\pm0.45$, the largest Gain@B of $5.72\%$, and the smallest Gap@O of $22.00\%$ among the evaluated representations.

ProbabilityMass uses 40 features, compared with 88 in the Full semantic representation, yet achieves an AvgAcc that is $0.78\%$  higher. Reducing the representation to 16 hard semantic judgments results in a further loss of information, with Hard achieving an AvgAcc of $0.88\%$ below ProbabilityMass. The remaining representations, i.e., GTE-Qwen2 embeddings, DomainOnly, and TF-IDF, also achieve competitive results, although none surpasses ProbabilityMass in AvgAcc. These differences are relatively small, but they show that a representation of explicit semantic features can perform well against larger dense embeddings and representations based on coarse domain or lexical information.

Across the five grouped splits, Best Single achieves an AvgAcc of $68.81\pm0.34$, while Dataset Oracle and Instance Oracle reach $73.94\pm0.51$ and $91.91\pm0.43$, respectively. ProbabilityMass exceeds Best Single by $3.27$ percentage points and remains $1.86$ percentage points below Dataset Oracle. The relatively small gap suggests that a substantial part of the achievable routing performance is associated with regularities at the dataset level. However, the larger difference between ProbabilityMass and Instance Oracle indicates that considerable instance-level opportunities remain that the current router does not capture.

\paragraph{Comparison with published routers.}
We also compare SeLMRoute with the performance results reported in LLMRouterBench~\cite{li2026llmrouterbench}. Table~\ref{tab:published_comparison} includes the published
AvgAcc values for RouterDC, GraphRouter, EmbedLLM, MODEL-SAT, and Avengers, alongside our results. SeLMRoute achieves an AvgAcc of $72.08\pm0.45$, which is above the reported results of these methods.

Note, however, that LLMRouterBench reports the published routers using its official prompt-grouped 30\% test protocol and an 11,480-prompt benchmark. Our internal protocol groups normalized router inputs and uses the 11,481-record frozen bundle described above. We therefore do not use the published values for paired statistical claims.

\begin{table}[t]
\centering
\small
\caption{AvgAcc compared with published LLMRouterBench performance-oriented results. Published methods use the benchmark protocol reported in ~\cite{li2026llmrouterbench}; SeLMRoute uses the duplicate-query-safe grouped protocol.}
\label{tab:published_comparison}
\begin{tabular}{lc}
\toprule
\textbf{Router} & \textbf{AvgAcc} \\
\midrule
RouterDC & 61.33 \\
GraphRouter & 70.29 \\
EmbedLLM & 71.24 \\
MODEL-SAT & 71.88 \\
Avengers & 71.94 \\
\textbf{SeLMRoute ProbabilityMass} & $\mathbf{72.08\pm0.45}$ \\
\midrule
Published Dataset Oracle & 73.10 \\
\bottomrule
\end{tabular}
\end{table}

The results place SeLMRoute within the performance range of the leading published routers, despite using a 40-dimensional representation of interpretable semantic evidence. The proximity of these methods to the Dataset Oracle is also consistent with LLMRouterBench's observation that dataset-level regularities account for a substantial part of routing performance.

\subsection{Probabilistic, Full, and Hard Semantic States}
\label{sec:representation_ablation}

We examine how the representation of semantic evidence affects routing performance. In particular, we compare three ways of representing the same sixteen semantic judgments: i) retaining their raw probability distributions (ProbabilityMass), ii) extending them with derived statistics (Full), or reducing each judgment to its most likely outcome (Hard). This comparison allows us to investigate whether preserving the probability distribution provides useful information to the downstream performance learner.

In the grouped five-fold OOF evaluation, ProbabilityMass achieves an AvgAcc of $72.64\%$, compared with $71.81\%$ for Hard, corresponding to a difference of $0.83\%$. The clustered 95\% bootstrap confidence interval is $[-0.174,\;1.837]$ percentage points, and the two-sided paired permutation test yields $p=0.1099$.
Although ProbabilityMass achieves higher accuracy, the confidence interval includes zero and the difference is not statistically significant at the 5\% level. The results therefore suggest a possible benefit from preserving probability mass, but do not establish that it consistently outperforms hard semantic judgments.

\begin{figure}[t] \centering \includegraphics[width=\columnwidth]{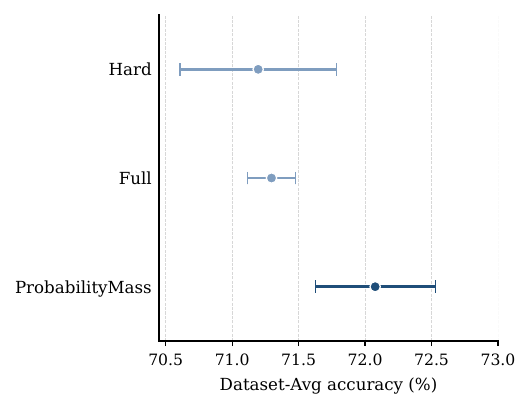} 
    \caption{Representation ablation. ProbabilityMass contains the 40 raw semantic probability features. Full contains 88 fine-grained semantic and derived features. Hard contains one discrete value per semantic probe.} \label{fig:representation_ablation}
\end{figure}

Figure~\ref{fig:representation_ablation} provides the corresponding comparison across the five grouped train-test splits. ProbabilityMass achieves the highest AvgAcc while using 40 features. The Full representation expands the semantic state to 88 features by adding derived quantities such as entropy, confidence, expected score, and distributional spread, but does not improve routing performance. Hard reduces the representation to sixteen discrete judgments and also achieves lower accuracy.

The results indicate that increasing the number of features does not necessarily improve routing performance. ProbabilityMass preserves the complete distribution associated with each semantic judgment without adding further quantities derived from the same distribution. Hard removes this information by retaining only the most likely outcome, while Full adds additional summary statistics. In our experiments, the raw probability representation provides the strongest performance while remaining substantially more compact than Full. The paired OOF comparison with Hard, however, shows that further evidence is required before concluding that this advantage generalizes beyond the current benchmark.

\subsection{Sensitivity to the Performance Learner}
\label{sec:router_ablation} 
\begin{figure}[t] 
\centering 
\includegraphics[width=\columnwidth]{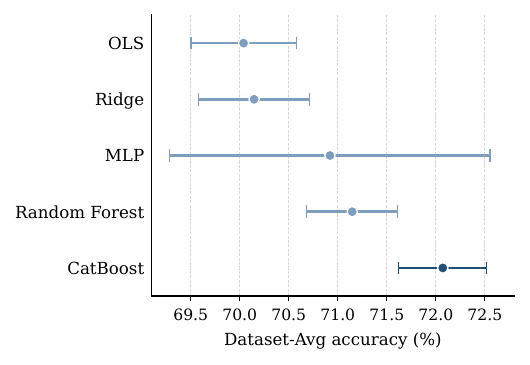} 
\caption{Performance-learner ablation with the ProbabilityMass semantic state held fixed.} \label{fig:router_ablation} 
\end{figure}

SeLMRoute does not require CatBoost as part of its semantic definition. In this experiment, we keep the ProbabilityMass representation fixed and replace the downstream performance learner. Figure~\ref{fig:router_ablation} compares CatBoost with Random Forest, a multilayer perceptron (MLP), Ridge regression, and ordinary least squares (OLS). CatBoost obtains the strongest result, while the difference between nonlinear and linear predictors suggests that the semantic variables contain useful information whose relationship with candidate performance is not purely additive.

Across the five grouped splits, CatBoost achieves $72.08\pm0.45$ AvgAcc, followed by Random Forest at $71.15\pm0.46$, MLP at $70.92\pm1.64$, Ridge at $70.15\pm0.57$, and OLS at $70.04\pm0.54$.

A simple example illustrates why such interactions may be useful. High code reasoning alone does not determine which model should receive a request. A code task with high exactness, high constraint density, and high context integration can favor a different model from a short code-completion request. A nonlinear performance learner can represent such interactions while the meaning of each semantic dimension remains unchanged.

\subsection{Does the Intermediate Semantic State Matter?}
\label{sec:direct_ablation}

A central design choice in SeLMRoute is to separate semantic evidence extraction from model selection. Instead of asking the decision model to select a candidate directly, we use it to describe the requirements of each query and train a separate performance learner to predict how the available models will perform. We examine the value of this separation through JEV Direct, a baseline in which JEV makes the routing decision without an intermediate semantic representation or a separate performance regressor.

In JEV-Direct, the decision model receives the input query along with anonymized capability profiles of the candidate models. The profiles are constructed exclusively from training-fold outcomes, with model identities hidden and no access to test outcomes. The decision model is asked to select the most suitable candidate via a single typed decision. Under the five-fold direct-routing experiment, JEV-Direct reaches $68.78\%$ AvgAcc, while SeLMRoute ProbabilityMass reaches $72.57\%$ on the same original fold assignment. The paired difference is $3.79\%$ in favor of SeLMRoute, with a 95\% interval of $[2.69,4.95]$ percentage points.

The difference supports the decomposition defined in Equation~\ref{eq:SeLMRoute_factorization}. A typed decision model can make a model choice directly, but measured candidate outcomes appear to be used more effectively when the decision model first produces reusable semantic evidence and a separate learner maps that evidence to model performance.

\subsection{Transfer Across Decision Models}
\label{sec:backend_transfer}

SeLMRoute is designed to operate independently of the decision model used for semantic extraction. To evaluate this property, we replace JEV with Laya, an open-weight typed decision model, while keeping the sixteen semantic probes, the 40-dimensional ProbabilityMass representation, and the downstream CatBoost routing procedure unchanged. Both configurations are evaluated using the same grouped five-fold OOF protocol, allowing us to examine the effect of changing the semantic extractor.

\begin{figure}[t] \centering \includegraphics[width=\columnwidth]{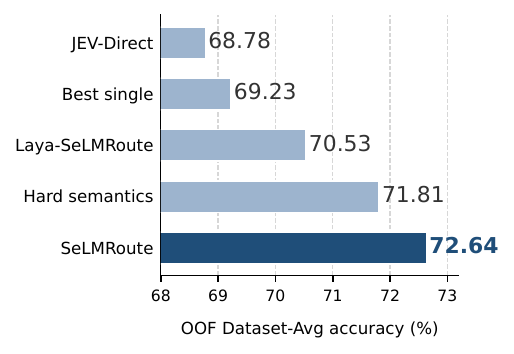} 
    \caption{Decision model comparison. Best Single is the strongest fixed candidate. JEV-SeLMRoute and Laya-SeLMRoute use the same semantic schema and downstream performance learner under the grouped five-fold OOF protocol.}
    \label{fig:architecture_comparison} 
\end{figure}

As presented in Figure~\ref{fig:architecture_comparison}, Laya-SeLMRoute achieves an AvgAcc of $70.53\%$, compared with $72.64\%$ for JEV-SeLMRoute and $69.23\%$ for Best Single. Thus, the open-weight backend maintains an advantage over the strongest fixed candidate, although its performance remains below that of JEV. The paired comparison between JEV and Laya yields a macro difference of $2.10$ percentage points, with a clustered 95\% bootstrap confidence interval of $[0.940,\;3.255]$ percentage points and a two-sided permutation test result of $p=0.00035$. The paired difference is statistically significant under the grouped evaluation protocol.

This finding shows that the semantic-state architecture can be used with different decision models without modifying the downstream routing procedure. However, preserving the same probe schema and representation does not guarantee equivalent routing performance. The difference between JEV and Laya suggests that the quality of the extracted semantic evidence has an important role in the accuracy of the resulting router. The open-weight experiment therefore supports the portability of SeLMRoute while highlighting the importance of the semantic extraction backend.

\subsection{Reducing the Semantic State}
\label{sec:probe_reduction}

Semantic extraction accounts for a substantial part of SeLMRoute's inference overhead, since each query must be evaluated against sixteen semantic probes. A natural question is whether fewer semantic questions can retain the same routing quality.

\begin{figure}[t]
    \centering
    \includegraphics[width=\columnwidth]{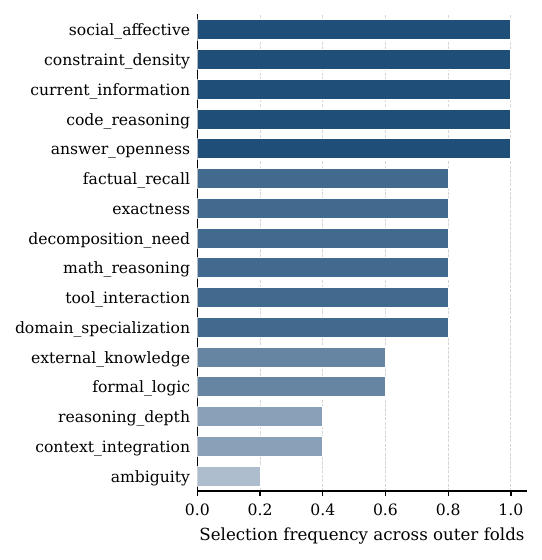}
    \caption{Probe-selection stability in the nested Lite-12 experiment. Each outer fold selects twelve probes using only its inner training data.}
    \label{fig:nested_probe_stability}
\end{figure}

We evaluate a configuration containing twelve of the original sixteen probes. A direct selection of the best probes on the full benchmark would leak test information into the semantic design. We instead use a nested protocol, where each outer training fold is divided into three inner folds. Probe importance is estimated exclusively inside the outer training partition, twelve probes are selected, and the resulting state is evaluated on the outer fold. The selection procedure is repeated independently for each of the five outer folds. Figure~\ref{fig:nested_probe_stability} shows the results of the nested experiment.

The full sixteen-probe ProbabilityMass router achieves an AvgAcc of $72.64\%$, while the reduced twelve-probe configuration reaches $72.49\%$. The observed difference is $-0.145$ percentage points, with a clustered 95\% paired bootstrap confidence interval of $[-0.903,\;0.614]$ percentage points. The two-sided paired permutation test yields $p=0.706$, providing no statistically significant evidence of a difference between the two configurations.

We additionally evaluate whether the twelve-probe configuration is noninferior to the full representation, using a predefined margin of $0.5$ percentage points. Although the observed performance difference falls within this margin, the one-sided 95\% lower confidence bound is
$-0.783$ percentage points. Noninferiority therefore cannot be established, as the interval does not exclude performance losses greater than the predefined margin.

The reduced configuration contains an average of 28.2 probability-mass features across the outer folds, compared with 40 in the full representation. The feature count varies because binary probes contribute one feature, whereas four-level Score probes contribute four. Reducing the number of probes decreases semantic input tokens by approximately $34\%$ and p95 semantic extraction latency by approximately $18\%$.

These results indicate that the semantic state can be compressed with a relatively small observed loss in routing accuracy and a measurable reduction in extraction overhead. However, the uncertainty around the performance difference prevents us from establishing that the reduced configuration retains the accuracy of the full sixteen-probe router within the predefined noninferiority margin.

\subsection{Which Semantic Judgments Matter?}
\label{sec:probe_importance}

We further investigate the contribution of individual semantic probes through a leave-one-probe-out ablation. Starting from the full sixteen-probe ProbabilityMass representation, we remove one probe at a time and retrain the downstream learner. Each configuration is evaluated using the same five grouped train-test splits as the main experiment. The performance difference is measured relative to the complete ProbabilityMass representation, allowing us to examine how removing each semantic judgment affects routing accuracy. Table~\ref{tab:probe_ablation} reports the change relative to the complete sixteen-probe ProbabilityMass representation. A positive value indicates that removing the probe decreases AvgAcc.

\begin{table}[t]
\centering
\small
\caption{Leave-one-probe-out results under grouped evaluation. Positive drop indicates lower AvgAcc after removing the probe.}
\label{tab:probe_ablation}
\begin{tabular}{lc}
\toprule
\textbf{Removed probe} & \textbf{Drop (pp)} \\
\midrule
Constraint density & +0.604 \\
Context integration & +0.511 \\
Current information & +0.487 \\
Tool interaction & +0.314 \\
Formal logic & +0.212 \\
Factual recall & +0.154 \\
Exactness & +0.128 \\
Reasoning depth & +0.118 \\
External knowledge & +0.064 \\
\midrule
Decomposition need & -0.028 \\
Social-affective & -0.072 \\
Code reasoning & -0.174 \\
Math reasoning & -0.179 \\
Ambiguity & -0.184 \\
Domain specialization & -0.190 \\
\bottomrule
\end{tabular}
\end{table}

The strongest reductions occur after removing \textit{constraint density}, \textit{context integration}, and \textit{current information}. The first two probes describe structural requirements that cut across benchmark domains. A mathematical problem, a programming task, and a factual question can all require the simultaneous satisfaction of many constraints. Their contribution is therefore consistent with the motivation of SeLMRoute, that routing-relevant information does not reduce to identifying whether a query belongs to mathematics, code, or another coarse category.

The result for \textit{current information} provides a different type of signal. Models can differ in how well they handle requests whose correctness depends on recent information, even when the broader subject of the query is familiar. Its positive marginal contribution shows that such requirements can help distinguish candidate behavior in the evaluated pool.

Several apparently intuitive probes show a negative leave-one-out drop. Removing \textit{domain specialization}, \textit{ambiguity}, \textit{math reasoning}, or \textit{code reasoning} slightly increases the five-seed mean. Nevertheless, the results do not imply that these concepts are irrelevant to routing. Leave-one-out analysis measures the \emph{marginal} value of one probe while all remaining semantic variables are present. A code-related requirement, for example, can also influence exactness, decomposition need, constraint density, external knowledge, and context integration. A dedicated code probe can therefore provide little unique information once those correlated signals are already available.

Small negative drops can additionally arise from estimation noise or from an extra feature encouraging the regressor to fit patterns that do not generalize across grouped splits. The magnitude of the negative changes is small, with the largest being $0.19\%$.

\subsection{Generalization Under Distribution Shift}
\label{sec:ood}

The preceding experiments evaluate SeLMRoute under in-distribution conditions, where all benchmark datasets contribute examples to both training and testing. In such a setting, the router can learn regularities associated with individual datasets and broader task domains. To examine how well the semantic representation generalizes beyond these conditions, we consider two out-of-distribution (OOD) evaluation protocols, Dataset-OOD and Domain-OOD.

\paragraph{Dataset OOD.}
In the first experiment, we hold out one complete dataset at a time, train the router on the remaining datasets, and evaluate it exclusively on the unseen dataset. This protocol examines whether the router can generalize to new benchmarks whose task requirements may still be represented in the training data.

ProbabilityMass achieves an AvgAcc of $67.02\%$, while Full reaches $67.27\%$, GTE-Qwen2 $67.29\%$, TF-IDF $66.68\%$, and DomainOnly $65.94\%$. All representations show lower accuracy than in the in-distribution evaluation, indicating that routing queries from an unseen dataset is more challenging than routing additional examples from datasets already represented during training.

The semantic representations remain competitive under this form of distribution shift. GTE-Qwen2 obtains the highest point estimate at $67.29\%$, followed closely by Full at $67.27\%$ and ProbabilityMass at $67.02\%$. The difference between GTE-Qwen2 and Full is only $0.02\%$, while Full exceeds ProbabilityMass by $0.25\%$. We therefore do not interpret the ordering as evidence for a clear advantage of one representation under Dataset-OOD. Instead, the results show that both explicit semantic evidence and a general dense representation can transfer to previously unseen datasets.

The performance of the semantic representations suggests that several routing-relevant properties recur across datasets. Queries from different benchmarks can require similar levels of reasoning depth, exactness, context integration, or constraint density even when their subjects and evaluation procedures differ. Such shared properties allow the performance learner to reuse relationships observed during training when processing queries from a previously unseen dataset.

\begin{figure}[t] \centering \includegraphics[width=\columnwidth]{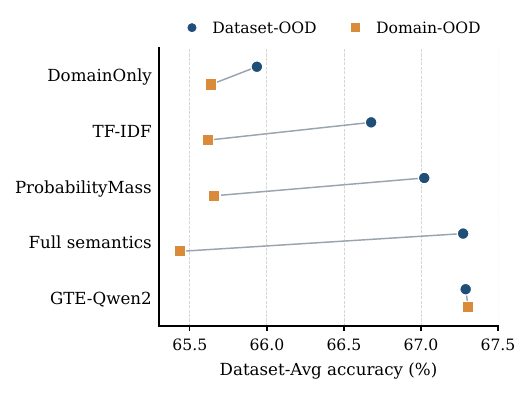}
    \caption{Generalization under distribution shift. Dataset-OOD holds out one dataset at a time, while Domain-OOD excludes all datasets belonging to one task domain.} 
    \label{fig:ood_generalization} 
\end{figure}

\paragraph{Domain-OOD.}
The second experiment considers a more challenging form of distribution shift by excluding an entire task domain from training. We evaluate five held-out domains: mathematics, code, logic, knowledge, and affective tasks. For each domain, the router is trained on all datasets
belonging to the remaining domains and evaluated on the excluded domain.
GTE-Qwen2 achieves the highest AvgAcc under Domain-OOD at $67.31\%$. ProbabilityMass reaches $65.66\%$, followed closely by DomainOnly at $65.64\%$ and TF-IDF at $65.62\%$. Full obtains $65.44\%$. Figure~\ref{fig:ood_generalization} summarizes the results of both OOD protocols.

The Domain-OOD experiment presents a more difficult form of generalization than holding out a single dataset. The semantic extractor can still describe queries from the excluded domain, but the downstream performance learner has never observed how the candidate models behave in that region of the task space. For example, when mathematics is excluded from training, a query can still receive high probability mass for mathematical reasoning, exactness, and reasoning depth. The missing information concerns how each candidate model performs for that combination of requirements.

GTE-Qwen2 shows a clear point advantage under this protocol, exceeding ProbabilityMass by approximately $1.65\%$. The remaining representations are considerably closer. Specifically, ProbabilityMass, DomainOnly, and TF-IDF all obtain AvgAcc values around $65.6\%$, while Full reaches $65.44\%$. The result differs from Dataset-OOD, where GTE-Qwen2, Full, and ProbabilityMass remain within approximately $0.27\%$  of one another.

One interpretation is that the dense GTE-Qwen2 representation retains latent information that is useful when the performance learner must extrapolate to an entirely unseen task domain. The explicit semantic state compresses each request into a predefined set of routing-relevant properties. Such compression is effective when the new dataset recombines requirements already represented during training, but it cannot recover empirical relationships between a completely unseen domain and candidate-model performance.

The two OOD experiments expose different transfer regimes. Dataset-OOD shows that semantic requirements can transfer across benchmark boundaries when related task characteristics remain represented in training. Domain-OOD shows a stronger limitation, i.e., describing an unfamiliar query through interpretable semantic properties does not remove the need for evidence about how candidate models behave in previously unseen regions of the task space.

\begin{table*}[t]
    \centering
    \small
    \caption{Grouped performance-cost evaluation. Monetary values refer to candidate model inference costs on the held-out test partition. Strict CostSave is reported only when the validation-selected $\theta^{\dagger}$ also satisfies the test accuracy requirement. N/A indicates that no configuration qualified under the strict protocol.}
    \label{tab:cost_results}
    \begin{tabular}{rcccccc}
    \toprule
    \textbf{Seed} &
    \textbf{GPT-5 AvgAcc} &
    \textbf{SeLMRoute AvgAcc} &
    \textbf{PerfGain} &
    \textbf{GPT-5 Cost} &
    \textbf{$\theta^{*}$ Cost} &
    \textbf{Strict CostSave} \\
    \midrule
    42   & 65.63 & 67.85 & +3.37\% & \$125.29 & \$133.30 & N/A \\
    3407 & 65.91 & 68.93 & +4.58\% & \$127.96 & \$128.09 & -0.10\% \\
    0    & 64.38 & 66.92 & +3.94\% & \$122.68 & \$112.83 & N/A \\
    1    & 64.92 & 65.48 & +0.87\% & \$127.72 & \$108.27 & N/A \\
    2    & 66.78 & 67.12 & +0.52\% & \$122.23 & \$124.16 & -1.58\% \\
    \midrule
    \multicolumn{3}{r}{\textbf{Mean PerfGain}} &
    $\mathbf{+2.66\pm1.85\%}$ &
    \multicolumn{3}{c}{} \\
    \bottomrule
\end{tabular}
\end{table*}

Taken together, the two experiments show that the semantic representation remains competitive under both forms of distribution shift. Dataset-OOD results indicate that semantic requirements learned across different benchmarks can support routing to unseen datasets. Domain-OOD presents a further challenge, that describing a query through reusable semantic properties does not eliminate the need for empirical evidence about candidate model performance. Addressing this limitation may require additional calibration or cold-start mechanisms.

\subsection{Semantic Uncertainty and Selective Routing}
\label{sec:uncertainty}

The probabilistic semantic state also provides a query-level measure of uncertainty. We examine whether this signal can identify queries for which the router is more likely to make an incorrect model selection. For each query, we compute the mean normalized entropy of the semantic probe distributions defined in Equation~\ref{eq:semantic_entropy}. We compare this quantity with the confidence margin of the downstream performance learner, defined from the separation between its highest-ranked candidate predictions. Both signals are evaluated using the same grouped OOF predictions used in the paired routing experiments.

We first evaluate how well each uncertainty measure ranks routing errors through the area under the risk-coverage curve (AURC). Queries are ordered from lower to higher confidence, progressively removed, and the routing error is recomputed over the retained subset. Lower AURC indicates that errors are concentrated more effectively among the queries rejected first. Semantic entropy obtains an AURC of $0.0901$, compared with $0.0974$ for the downstream router margin. Under the in-distribution grouped OOF protocol, uncertainty in the semantic representation therefore provides a slightly better ranking of difficult routing cases than the confidence margin derived from the final candidate predictions.

We next test whether this uncertainty can be converted into a rejection rule that transfers beyond the calibration distribution. For each of five Domain-OOD splits, the entropy threshold is selected using only in-distribution calibration data and then kept fixed when evaluating the held-out domain. Queries whose semantic entropy exceeds the calibrated threshold are rejected, while the remaining queries are routed normally. Across the five held-out-domain evaluations, selective routing achieves an AvgAcc of $65.38\pm0.60\%$ at $96.82\pm0.92\%$ coverage. Routing all queries without rejection reaches $65.32\pm0.60\%$ under the same Domain-OOD protocol.

Semantic entropy is useful for \emph{ranking} difficult queries within the distribution on which uncertainty is observed, but a fixed threshold calibrated on familiar data does not produce a meaningful accuracy gain after transfer to an unseen domain. The result is consistent with the role of the semantic state, i.e., entropy measures uncertainty in how the query is characterized by the semantic extractor, whereas routing uncertainty also depends on how much evidence the performance learner has about candidate behavior in that region of the semantic space. A query can therefore receive a confident semantic description while still lying in a region where candidate performance is poorly supported by the training data.

\subsection{Performance-Cost Routing}
\label{sec:cost_results}

We investigate whether the ProbabilityMass representation can also support cost-aware routing across a different candidate pool. For this experiment, we use the LLMRouterBench performance-cost setting, which contains 13 flagship models and 10 datasets. Our benchmark bundle includes 12,446 queries and 161,520 observed candidate outcomes. Since candidate coverage is incomplete, we train model-specific quality and cost predictors using the available observations, without imputing missing scores or costs.

In this experiment, cost refers to the monetary API inference cost of the selected candidate model, measured in dollars. LLMRouterBench records the number of input and
output tokens together with the corresponding model-specific
API prices~\cite{li2026llmrouterbench}. The cost of a single
model invocation is calculated as:
\begin{equation*}
    c_m(x)
    =
    \frac{
        t^{\mathrm{in}}_m(x)p^{\mathrm{in}}_m
        +
        t^{\mathrm{out}}_m(x)p^{\mathrm{out}}_m
    }{10^6},
\end{equation*}

where $t^{\mathrm{in}}_m(x)$ and $t^{\mathrm{out}}_m(x)$ denote the input and output token counts, and $p^{\mathrm{in}}_m$ and $p^{\mathrm{out}}_m$ are the corresponding prices per million tokens. For tasks involving multiple model invocations, the recorded cost is the sum of the individual invocation costs. The total candidate model inference cost of a routing
policy $r$ is $\operatorname{Cost}(r) = \sum_i c_{r(x_i)}(x_i)$.

We use GPT-5 as a fixed reference model across all experimental splits. In the LLMRouterBench flagship pool, GPT-5 is the Best Single model, achieving the highest macro-average accuracy across the ten performance-cost datasets while providing complete benchmark coverage. The published benchmark also uses GPT-5 as its reference for PerfGain and CostSave~\cite{li2026llmrouterbench}.

We use a grouped 70/30 train-test split, reserving 20\% of the outer training partition for validation. The routing parameter $\lambda$, defined in Equation~\ref{eq:cost_utility}, controls the trade-off between predicted quality and inference cost and its value
is selected exclusively using the validation partition.

We consider two operating points. The first, $\theta^{*}$, maximizes validation AvgAcc and is used to measure the performance improvement achieved by routing. The second, $\theta^{\dagger}$, is the least-cost validation configuration whose AvgAcc is at least as high as that of GPT-5 on the same validation partition. The latter operating point is used to evaluate potential monetary savings without sacrificing the reference model's accuracy.

After selecting the operating points, we refit the quality and cost predictors on the complete outer training partition. The selected values of $\lambda$ remain fixed during evaluation on the held-out test set.

PerfGain is calculated from the test performance of $\theta^{*}$ relative to GPT-5. Strict CostSave is reported only when a qualifying $\theta^{\dagger}$ exists on validation and also maintains or exceeds GPT-5 accuracy on the test partition. When either condition is not satisfied, strict CostSave is reported as unavailable.

This distinction prevents us from selecting a cheaper operating point retrospectively based on favorable test outcomes. In particular, a configuration that achieves higher accuracy and lower cost on the test set does not qualify for strict CostSave unless it also satisfies the predefined validation criterion. Table~\ref{tab:cost_results} presents the cost results.

PerfGain is positive in all five grouped splits. The improvements range from $+0.52\%$ to $+4.58\%$, with a mean of $+2.66\pm1.85\%$. SeLMRoute therefore improves test performance relative to GPT-5 under every evaluated split of the flagship-model benchmark.

The monetary results are less consistent. Positive strict CostSave is not achieved in any of the five splits. Only seeds 3407 and 2 produce validation-selected $\theta^{\dagger}$ configurations that also satisfy the held-out accuracy requirement. Their respective CostSave values are $-0.10\%$ and $-1.58\%$, indicating that the qualifying configurations are slightly more expensive than GPT-5.

Seeds 0 and 1 illustrate the importance of separating performance-oriented routing from validation-qualified cost savings. In both cases, the selected $\theta^{*}$ achieves higher test accuracy than GPT-5 at a lower observed candidate model cost. However, no
$\theta^{\dagger}$ satisfies the validation-stage accuracy requirement. These outcomes therefore do not qualify as strict CostSave under our protocol, even though their final test results are favorable.

Overall, the experiment provides evidence that SeLMRoute can improve candidate selection in the flagship-model setting. The same results do not establish reliable monetary savings. Although some performance-oriented configurations are also cheaper on the test set, the current validation and utility selection procedure does not consistently identify a lower-cost operating point that preserves GPT-5 accuracy before the test outcomes are observed.

\subsection{Dollar Cost of Semantic Extraction}
\label{sec:semantic_dollar_cost}

The preceding performance-cost experiment measures candidate model inference costs but excludes the additional API calls required to generate SeLMRoute's semantic representation. We therefore estimate the monetary overhead of semantic extraction separately, using the recorded token consumption of our performance-oriented experiments.

At the time of evaluation, JEV was priced at \$0.042 per million input tokens, with no charge for output tokens~\cite{almeida2026jev}. Extracting the complete sixteen-probe semantic state requires an average of 1,606.7 input tokens per query. Across the 11,481-query performance-oriented benchmark, the recorded consumption is 18,446,536 input tokens.
The estimated extraction cost is therefore $18{,}446{,}536 \times \frac{\$0.042}{10^6}  = \$0.775$, corresponding to $~\$6.75\times10^{-5}$ per query, or \$0.0675 per 1,000 queries.

We also estimate the cost of the reduced twelve-probe configuration evaluated in Section~\ref{sec:probe_reduction}. Lite-12 requires an average of 1,062.7 input tokens per query, with a recorded total of 12,200,872 tokens across the same benchmark. At the same API rate, its estimated monetary cost is \$0.512, equivalent to approximately \$0.0446 per 1,000 queries. This represents a reduction of approximately $33.9\%$ in semantic-extraction input cost.

Table~\ref{tab:semantic_cost} summarizes the estimated costs of both configurations. These estimates are reported separately from the CostSave results because the token measurements come from the performance-oriented benchmark, whereas cost-aware routing is evaluated on a different pool of 12,446 queries. Directly adding the estimated extraction expenditure to the performance-cost results would combine measurements from different query populations.

Nevertheless, the estimates provide an indication of the additional monetary overhead introduced by the semantic stage. At the evaluation-time JEV price, the API charge for extracting the semantic representation is small relative to the recorded cost of flagship-model inference. Reducing the number of probes lowers this expenditure further.

The monetary estimate does not capture the complete operational overhead of semantic
extraction. In particular, the additional decision-model invocation introduces latency, which may affect the suitability of SeLMRoute for time-sensitive deployments. We examine this aspect separately in Section~\ref{sec:systems}.
\begin{table}[t]
    \centering
    \small
    \caption{Estimated semantic extraction cost on the 11,481-query performance-oriented benchmark. PM-16 uses all sixteen semantic probes and the 40-feature ProbabilityMass representation. Candidate model inference costs are excluded.}
    \label{tab:semantic_cost}
    \begin{tabular}{lrrrr}
    \toprule
    \textbf{Config} &
    \textbf{Mean input} &
    \textbf{Total input} &
    \textbf{\$/1K queries} &
    \textbf{Total \$} \\
    \midrule
    PM-16 & 1,606.7 & 18.45M & 0.0675 & 0.775 \\
    Lite-12 & 1,062.7 & 12.20M & 0.0446 & 0.512 \\
    \bottomrule
    \end{tabular}
\end{table}

\subsection{System Overhead}
\label{sec:systems}

As a final experiment, we examine the inference overhead introduced by SeLMRoute, separating semantic extraction from downstream performance prediction. The measurements show that semantic extraction accounts for most of the additional latency, while the
CatBoost performance learner introduces comparatively little overhead.

\paragraph{Semantic extraction.}
The complete sixteen-probe JEV configuration consumes approximately 1,607 input tokens per query on average, with a 95th-percentile (p95) token count of 1,919. Its mean observed extraction latency is approximately $350$ ms, increasing to $481$ ms at p95. These measurements include remote API processing and network communication. They therefore represent the observed latency of the semantic extraction stage and not the isolated
execution time of the underlying decision model.

\begin{figure}[t]
    \centering
    \includegraphics[width=\columnwidth]{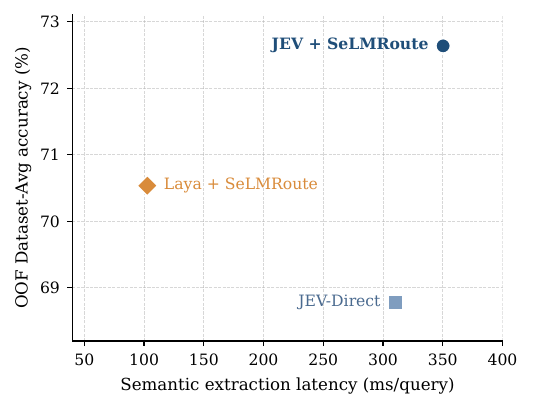}
    \caption{Routing accuracy and observed
    semantic-extraction latency for JEV-SeLMRoute
    and Laya-SeLMRoute. Accuracy is measured using
    grouped five-fold OOF evaluation. JEV latency
    includes remote API and network overhead,
    whereas Laya is measured locally. The latency
    measurements reflect the evaluated deployment
    environments and are not hardware-normalized
    model benchmarks.}
    \label{fig:systems_tradeoff}
\end{figure}

We also measure the open-weight Laya backend in our local execution environment. Laya records a mean extraction latency of $~103$ ms and a p95 latency of $~206$ ms. Figure~\ref{fig:systems_tradeoff} presents the observed extraction latency alongside the grouped OOF routing accuracy of both backends.

The measurements illustrate the accuracy-latency trade-off of the evaluated implementations, although they should not be interpreted as a direct comparison of model execution speed. JEV is accessed through a remote API, whereas Laya runs locally, and the two measurements therefore include different sources of overhead. As discussed in Section~\ref{sec:backend_transfer}, JEV-SeLMRoute also achieves higher routing accuracy than Laya-SeLMRoute under the shared semantic schema.

\paragraph{Downstream routing.}
Once the semantic state has been extracted, the CatBoost performance learner requires $~1.23$ ms to process a single query at batch size one. With batching, the processing time falls to $~0.043$ ms per query at batch size 32 and $~0.011$ ms at batch size 256. In all three configurations, the downstream
prediction stage is substantially faster than semantic extraction.

\paragraph{Effect of probe reduction.}
Reducing the semantic state provides an additional way to lower inference overhead. The twelve-probe
configuration decreases mean semantic input from 1,606.7 to 1,062.7 tokens and reduces p95 extraction
latency from approximately $~481$ ms to $~394$ ms. This corresponds to reductions of $~34\%$ in input tokens and $~18\%$ in p95 latency. The observed grouped OOF accuracy decreases by only $0.145$ percentage points, although, as discussed in Section~\ref{sec:probe_reduction}, statistical
noninferiority to the full configuration is not established.

The measurements suggest that semantic extraction acts as the main additional latency cost of SeLMRoute. The downstream performance learner is comparatively lightweight, and its processing time decreases further with batching. Although the estimated monetary cost of JEV extraction is low under the evaluated API pricing, its latency is more consequential for deployments requiring short response times. Further efficiency work should therefore focus on reducing semantic extraction overhead, including adaptive probe execution and faster or specialized decision backends.

\section{Conclusions and Future Work}
\label{sec:conclusion}
In this paper we introduced SeLMRoute, a framework that separates interpretable probabilistic semantic evidence from model performance learning and routing objectives. Across five grouped splits, SeLMRoute achieves $72.08\pm0.45\%$ AvgAcc, remaining competitive with established routing approaches. Our experiments demonstrate that the semantic interface can support different decision models and candidate pools, while preserving probability mass offers a compact alternative to larger representations. In the performance-cost setting, SeLMRoute achieves a mean PerfGain of $2.66\%$, although reliable monetary savings are not established.

Several open problems remain. Generalization to entirely unseen domains is still difficult, semantic extraction introduces measurable latency, and improved quality prediction does not necessarily translate into monetary savings. Another limitation arises when a new candidate LLM is added to the pool. Previously extracted semantic states remain reusable, but the performance learner has no prior evidence about the new model and therefore requires candidate-specific calibration.

Future work will investigate decision models specialized for semantic routing, including fine-tuning an open decision model directly on the typed semantic-analysis task used by SeLMRoute. We also plan to study calibration of the resulting probability distributions, joint treatment of semantic and routing uncertainty, and adaptive probe selection. Candidate cold start is another priority, with active calibration strategies that select a small but informative set of semantic regions for evaluating newly introduced models. Further work will examine online adaptation as candidate capabilities or prices change, and routing objectives that jointly account for model quality, candidate inference cost, semantic-extraction cost, and latency.

\bibliographystyle{IEEEtran}
\bibliography{references}

\end{document}